\documentclass{article}
\PassOptionsToPackage{numbers, sort&compress}{natbib}
\usepackage[dblblindworkshop, final]{neurips_2026}
\workshoptitle{Physical World AI: Geometry, Characteristics, and Multimodal Sensing}

\usepackage[utf8]{inputenc} 
\usepackage[T1]{fontenc}    
\usepackage{hyperref}       
\usepackage{url}            
\usepackage{booktabs}       
\usepackage{amsfonts}       
\usepackage{nicefrac}       
\usepackage{microtype}      
\usepackage{xcolor}         

\usepackage{graphicx}
\usepackage{capt-of}
\usepackage{tabularx}

\title{Separating Environment and Sensor Shift from Acquisition Protocol in Physical Vision Benchmarks}

\author{%
  Yuyang Yuan \\
  Electrical and Computer Engineering\\
  University of California, Los Angeles\\
  \texttt{yuanyuyang@g.ucla.edu} \\
  \And
  Jason Wu \thanks{Jason Wu is currently with Amazon. The work described in this paper was performed while he was at the University of California, Los Angeles, and is not associated with Amazon.}\\
  Electrical and Computer Engineering \\
  University of California, Los Angeles \\
  \texttt{jaysunwu@g.ucla.edu} \\
  \AND
  Mohammadali Khodabandehlou \\
  Electrical and Computer Engineering \\
  University of Southern California \\
  \texttt{mk40089@usc.edu} \\
  \And
  Jianpeng Yao \\
  Electrical and Computer Engineering \\
  University of California, Riverside \\
  \texttt{jyao073@ucr.edu} \\
  \And
  Hang Qiu\\
  Electrical and Computer Engineering \\
  University of California, Riverside \\
  \texttt{hangq@ucr.edu} \\
  \And
  Mani Srivastava\\
  Electrical and Computer Engineering \\
  University of California, Los Angeles \\
  \texttt{mbs@ucla.edu} \\
}

\begin{document}

\maketitle










\begin{abstract}

  In real-world deployments, vision systems must operate under varying environmental conditions, such as illumination changes. Recent work has explored adaptive sensing strategies that select camera parameters according to the current environment and downstream task to improve task performance. Reliable physical benchmarks are therefore essential for evaluating such strategies. We study ImageNet-ES-Diverse, a physical benchmark that evaluates the adaptive exposure control strategy \textit{Lens} by recapturing printed ImageNet images under varying illumination conditions. We construct four controlled acquisition pipelines that vary printed-image scale and camera zoom, and conduct a pilot study with directly captured physical objects. We find that the acquisition choices substantially change benchmark difficulty on all downstream baselines. Moreover, the average classification-accuracy advantage of \textit{Lens} over auto exposure observed on the original benchmark does not persist in either our collected pipelines or real-object pilot study. Oracle exposure selection nevertheless retains substantial gains, indicating remaining potential for sensor control. These findings highlight acquisition protocol as an important factor in physical benchmark evaluation and motivate explicit protocol reporting and evaluation across multiple acquisition setups when assessing adaptive-sensing benefits.
\end{abstract}



\section{Introduction}
The real-world deployment of vision algorithms in applications such as autonomous driving and robotics exposes sensing systems to diverse and continuously changing scene and environmental conditions. Under these conditions, downstream task performance depends not only on the vision algorithm itself, but also on how the scene is captured through the selected camera parameters~\cite{Onzon_2021_CVPR, Baek_2024_CVPR}. Recent work has explored adaptive sensing strategies that dynamically select sensor parameters according to the current environment or downstream objective~\cite{Baek_2024_CVPR,yan2025joca, zhang2024efficient}. Evaluation of such strategies therefore relies on benchmarks that systematically cover diverse combinations of environmental conditions and sensor parameters through digital post-processing, simulation, or real-world data capture~\cite{hendrycks2019benchmarking,dosovitskiy2017carla,savva2019habitat, Baek_2024_CVPR, baek2025adaptive}. Digital post-processing modifies observations only after image acquisition while simulated environments may not fully reproduce the physics of the scene, so neither approach fully reproduces the effects of physical image acquisition. Thus, recent works have constructed physical benchmarks involving real cameras subjected to sensor variations in a controlled environment~\cite{Baek_2024_CVPR,baek2025adaptive}.


One challenge associated with physical benchmark construction involves acquiring a large number of sensing targets for sufficient diversity among both classes and samples within a class, which can be costly and time-consuming. A common strategy involves physically reproducing the source image from another dataset (e.g., ImageNet) by printing or displaying the sample on a designated medium, and then recapturing it under different illumination conditions and camera parameter settings~\cite{Baek_2024_CVPR, baek2025adaptive}.  
However, this physical recapture protocol can induce unintentional side effects: the captured observation now depends not only on the intended environmental conditions and camera parameters, but also on the physical reproduction of the source image and how the medium on which it is displayed interacts with the camera capture process. This raises a fundamental question for physical vision benchmarks: do the captured observations reflect only the intended environment and camera variations, or a mixture of these variations and protocol-induced effects?

In this work, we use ImageNet-ES-Diverse~\cite{baek2025adaptive} (IED) as a case study to examine how acquisition choices affect classification performance and whether the reported Lens-over-Auto Exposure (AE) advantage persists across independently constructed acquisition setups. Based on the available description of its acquisition setup, we construct a reference recapture setup and compare it with several alternative recapture protocols while using the same source images, lighting configurations, and manual exposure grid across our four pipelines. Based on the four controlled acquisition pipelines that vary
printed-image scale and camera zoom, we find that the recapture protocol can substantially change downstream baseline performance, demonstrating that the physical benchmark difficulty is strongly protocol-dependent. More importantly, evaluating adaptive sensor control under different recapture protocols can substantially change its average measured advantage over conventional baselines, challenging previous conclusions drawn from a single acquisition protocol. We further examine this finding in a real-object pilot study, where the same overall trend persists without relying on printed-image recapture. These results suggest that conclusions about the effectiveness of adaptive sensing should be interpreted jointly with the physical acquisition protocol used to construct the benchmark.

\section{Background}
\subsection{Benchmark Construction for Sensor and Environmental Variations}

Existing benchmarks study vision systems under varying environmental conditions and camera configurations. One approach digitally transforms existing image datasets to approximate these variations~\cite{hendrycks2019benchmarking,michaelis2020benchmarkingrobustnessobjectdetection}, avoiding repeated physical data collection. However, because these transformations operate on already captured and processed images, they cannot recover information lost during the original image-formation process.

Another approach uses physics-based 3D simulators such as CARLA and Habitat~\cite{dosovitskiy2017carla,savva2019habitat}. TaCOS~\cite{yan2024tacostaskspecificcameraoptimization} uses simulated observations to jointly optimize camera parameters and downstream perception models, while JOCA~\cite{yan2025joca} uses CARLA under varying illumination to jointly optimize camera hardware, adaptive exposure control, and object detection. Unlike digital post-processing, simulation explicitly models image formation. However, simulated rendering and sensor pipelines remain approximations of real-world processes, introducing a potential sim-to-real gap.

To capture real image-formation effects, several benchmarks use physical recapture. ImageNet-ES~\cite{Baek_2024_CVPR} displays a subset of Tiny-ImageNet on a screen and recaptures the images under controlled lighting and camera settings. ImageNet-ES-Diverse~\cite{baek2025adaptive} follows a similar approach but prints source images on PVC banners. While improving physical realism, recapture may introduce protocol-induced shifts beyond the intended environmental and sensor variations.

\subsection{Acquisition Pipeline}

We use ImageNet-ES-Diverse~\cite{baek2025adaptive} as a case study. It selects 1,000 reference images from Tiny-ImageNet, each reproduced on an A4-sized PVC banner printed at 600 DPI. The banners are recaptured in a controlled darkroom using a fixed camera setup under varying illumination and combinations of ISO, shutter speed, and aperture. For each source image, a region of interest is manually defined on one capture and reused to crop all captures across environmental and camera settings. This produces cropped physical observations under controlled illumination and sensor-parameter variations.

\section{Methodology: Controlled Acquisition Pipeline}
\subsection{Reference Recapture Setup}
\begin{figure}[t]
\centering

\begin{minipage}[t]{0.44\linewidth}
    \vspace{0pt}
    \centering

    \includegraphics[width=\linewidth]{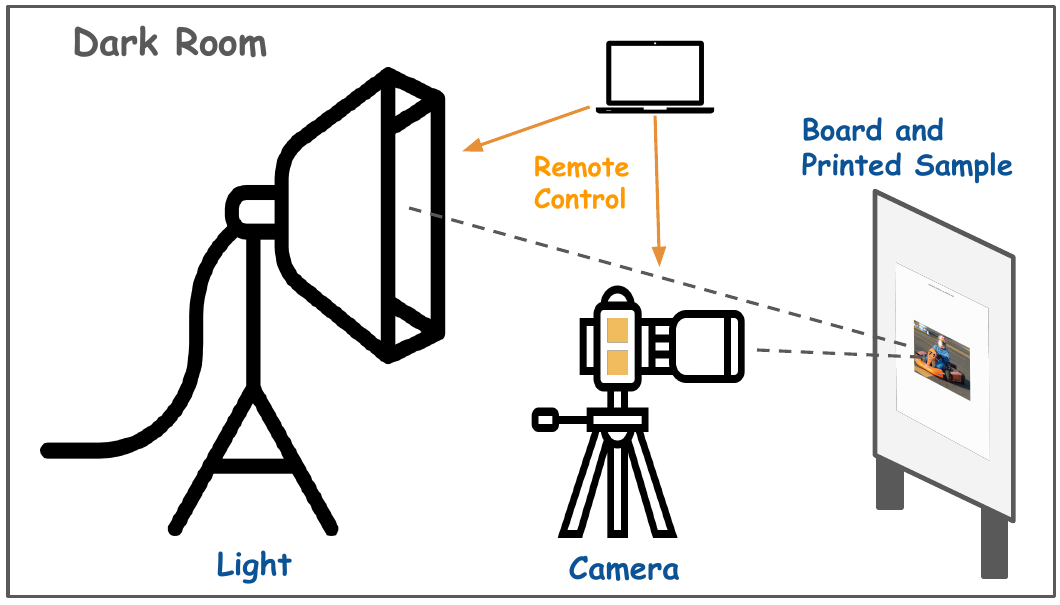}
    \caption{Physical acquisition setup used for printed-image recapture.}
    \label{fig:physical_setup}

    \vspace{8pt}
    \setlength{\abovecaptionskip}{0pt}
    \setlength{\belowcaptionskip}{7pt}

    \captionof{table}{Lighting setup and illumination conditions.}
    \label{tab:lighting}

    \small
    \setlength{\tabcolsep}{2pt}
    \begin{tabularx}{\linewidth}{
        @{}l*{5}{>{\centering\arraybackslash}X}@{}
    }
        \toprule
        & L1 & L2 & L3 & L4 & L5 \\
        \midrule
        Intensity (\%) & 1 & 20 & 50 & 70 & 100 \\
        \midrule
        Model
            & \multicolumn{5}{c@{}}{Amaran Halo 300X} \\
        CCT
            & \multicolumn{5}{c@{}}{5600 K} \\
        Modifier
            & \multicolumn{5}{c@{}}{Aputure 2 x 3ft softbox} \\
        \bottomrule
    \end{tabularx}
\end{minipage}\hfill
\begin{minipage}[t]{0.53\linewidth}
    \vspace{0pt}
    \centering

    \setlength{\abovecaptionskip}{0pt}
    \setlength{\belowcaptionskip}{7pt}

    \captionof{table}{Camera configurations used in our printed-image data
    \mbox{collection}.
    \textcolor{blue}{Blue} denotes settings not explicitly reported in
    ImageNet-ES-Diverse and specified in our setup;
    \textcolor{red}{red} denotes settings different from the original
    benchmark; and \textcolor{green!50!black}{green} denotes settings
    following the benchmark.}
    \label{tab:camera_settings}

    \small
    \setlength{\tabcolsep}{2pt}
    \begin{tabularx}{\linewidth}{
        @{}lc>{\centering\arraybackslash}X@{}
    }
        \toprule
        \multicolumn{3}{@{}l}{\textbf{Fixed Settings}} \\
        \midrule
        Camera body + lens
            & \multicolumn{2}{c@{}}{
                \textcolor{red}{Sony a7R VI (ILCE-7RM6)}} \\
            & \multicolumn{2}{c@{}}{
                \textcolor{red}{+ FE 24--70mm F2.8 GM II}} \\
        Focus mode
            & \multicolumn{2}{c@{}}{
                \textcolor{red}{Manual Focus}} \\
        White balance
            & \multicolumn{2}{c@{}}{
                \textcolor{blue}{Auto}} \\
        Image resolution
            & \multicolumn{2}{c@{}}{
                \textcolor{green!50!black}{3:2},
                \textcolor{red}{$6528\times4352$}} \\
        Metering mode
            & \multicolumn{2}{c@{}}{
                \textcolor{blue}{Multi}} \\
        Image quality
            & \multicolumn{2}{c@{}}{
                \textcolor{blue}{JPEG (Fine)}} \\
        \midrule
        \multicolumn{3}{@{}l}{\textbf{Exposure Settings}} \\
        \addlinespace
            & \textbf{AE} & \textbf{Manual} \\
        \midrule
        No. of shots
            & \textcolor{red}{3}
            & \textcolor{green!50!black}{27} \\
        ISO
            & \textcolor{green!50!black}{Auto}
            & \textcolor{green!50!black}{250, 2000, 16000} \\
        Aperture
            & \textcolor{green!50!black}{Auto}
            & \textcolor{green!50!black}{
                $f$/5.0, $f$/9.0, $f$/16} \\
        Shutter speed
            & \textcolor{green!50!black}{Auto}
            & \textcolor{green!50!black}{
                1/4\,s, 1/60\,s, 1/1000\,s} \\
        \bottomrule
    \end{tabularx}
\end{minipage}

\end{figure}

\begin{figure}[t]
    \centering
    \includegraphics[width=\linewidth]{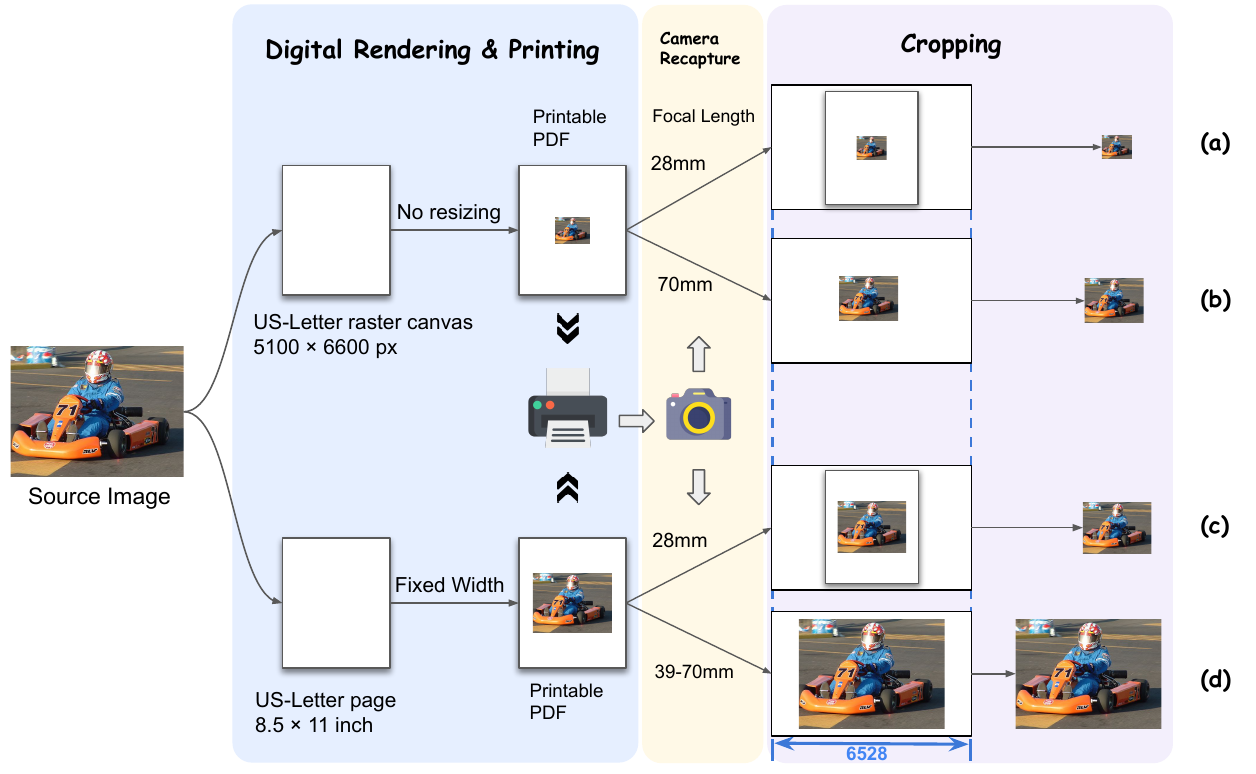}
    \caption{Printed-image acquisition pipeline and controlled variants. We consider four pipeline configurations: (a) original-resolution rendering without resizing, captured at 28 mm; (b) original-resolution rendering without resizing, captured at 70 mm; (c) fixed-width rendering, captured at 28 mm; and (d) fixed-width rendering with zoomed-in camera framing.}
    \label{fig:pipelines}
\end{figure}

We construct a reference recapture setup informed by the reported acquisition design of ImageNet-ES-Diverse \cite{baek2025adaptive}. As illustrated in Figure~\ref{fig:physical_setup}, the experiment is conducted in a dark room, with a controllable studio light, a tripod-mounted camera, and the printed sample mounted on a white board. During setup, we first level the camera in both roll and pitch using its built-in electronic level. We then align the edges of the printed image with the camera's on-screen reference lines and match the center of the frame with the center of the target sample as closely as possible. Once aligned, the physical positions of the light, camera, and target board are fixed throughout data collection. Each printed sample is mounted at the same predefined position on the board using push pins to ensure consistent placement. As part of our reference acquisition pipeline described in Figure ~\ref{fig:pipelines}, we deliberately set the camera-to-target distance to 21 cm to ensure that the full printed page approximately fills the vertical field of view at a focal length of 28\,mm.

To generate different physical conditions, we remotely control the illumination intensity and camera parameters from a laptop. We use the manual exposure grid reported in ImageNet-ES-Diverse \cite{baek2025adaptive}, with other acquisition settings detailed below. For illumination, we use a remotely controlled studio light and define five intensity levels spanning its controllable range, as summarized in Table~\ref{tab:lighting}. For each lighting condition, we capture three images in auto-exposure (AE) mode and 27 images in manual mode, using a $3\times3\times3$ sweep over ISO, aperture, and shutter speed (Table~\ref{tab:camera_settings}). Table~\ref{tab:camera_settings} distinguishes settings shared with ImageNet-ES-Diverse, modified settings, and settings specified in our setup. We use manual focus instead of auto focus as the camera and target are at fixed positions, providing more consistent focusing across captures and faster automated acquisition.

A less explicitly specified part of ImageNet-ES-Diverse is how each digital source image is converted into the physical printed target and subsequently into the final cropped observation. The paper reports the use of A4-sized PVC banners printed at 600 DPI and a camera capture resolution of $6240\times4160$, while the intermediate source-to-print generation procedure is not fully specified. Based on the reported information and statistics of the released dataset, we reconstruct a reference acquisition pipeline. The pipeline covers PDF construction, printing, camera recapturing and region of interest (ROI) cropping as illustrated in Figure~\ref{fig:pipelines}(a):

\begin{enumerate}
    \item Create a $5100\times6600$-pixel white canvas corresponding
    to a Letter-size page at 600 PPI: $(8.5,11)\,\mathrm{in}\times600\,\mathrm{pixels/in}=(5100,6600)\,\mathrm{pixels}$
    \item Place the corresponding reference image from the
    ImageNet-ES-Diverse source set at its original pixel resolution
    and aspect ratio at the center of the canvas, without resizing (1 source px = 1 canvas px),
    and export the page as a printable PDF.
    \item Print the generated page on Letter-size vinyl paper at
    600 DPI.
    \item Recapture the printed page with the focal length fixed at
    28\,mm. At the predefined camera-to-board distance, the full
    printed page approximately fills the vertical field of view,
    leaving only a small margin above and below the page.

    \item Repeat the capture under all lighting conditions and camera
    configurations described above.

    \item Manually determine the ROI for each sample and automatically
    reuse the same cropping coordinates across all lighting and camera
    configurations of that sample.
\end{enumerate}

Appendix~\ref{Appendix:resolution} reports the cropped-image dimensions for each of the 20 samples in IED and our four pipelines, allowing a direct per-sample comparison.

We apply the same acquisition procedure to 20 printed samples randomly selected from the 1,000 reference samples used in ImageNet-ES-Diverse.

\subsection{Controlled Pipeline Variations}

In the reference pipeline (a), the source image is placed without resizing on our 600-PPI canvas. Consequently, the physical size of the printed ROI can be small relative to the page, thus occupying only a small portion of the camera frame after recapture. To study how the acquisition protocol affects the resulting observation, we construct three controlled variants that increase the ROI occupancy through either camera framing, physical reproduction scale, or both.

Figure~\ref{fig:pipelines} pipeline (b) changes only the camera framing while keeping the printed target and all other acquisition settings identical to pipeline (a). Specifically, we increase the focal length from 28\,mm to 70\,mm, the maximum focal length of the lens, so that the printed ROI occupies a larger fraction of the captured frame. We ensure that the complete ROI remains within the camera field of view.

Figure~\ref{fig:pipelines} pipeline (c) instead changes the digital rendering process to increase the physical size of the source image on the printed page, while keeping the camera focal length at 28\,mm. We construct the printable target as follows:
\begin{enumerate}
    \item Create a US-Letter PDF page with a physical size of $8.5\times11$ inches. Unlike pipeline (a), the page is defined directly by its physical dimensions rather than by a raster canvas with a predefined PPI.
    
    \item Embed the corresponding reference image at a fixed physical width of $60\%$ of the page width
    ($0.6 \times 8.5\,\mathrm{in} = 5.1\,\mathrm{in}$),
    while preserving its aspect ratio and original pixel resolution. The source image is therefore not explicitly upsampled before being embedded into the PDF.
    
    \item Export the composed page as a printable PDF and follow the same printing, recapturing, and ROI-cropping procedure as in pipeline (a).
\end{enumerate}

This design enlarges the physical reproduction of the source image without first creating additional source pixels. The resampling required to realize the enlarged image occurs during print rasterization rather than through explicit pre-upsampling in our rendering pipeline. With the same 28\,mm camera setting, the larger physical target consequently occupies a larger region of the captured frame.

Finally, Figure~\ref{fig:pipelines} pipeline (d) combines the fixed-width rendering of pipeline (c) with zoomed-in camera framing. We increase the focal length while ensuring that the complete printed ROI remains inside the captured frame, thereby further increasing its occupancy in the camera frame.

\begin{figure}[t]
    \centering
    \includegraphics[width=0.7\linewidth]{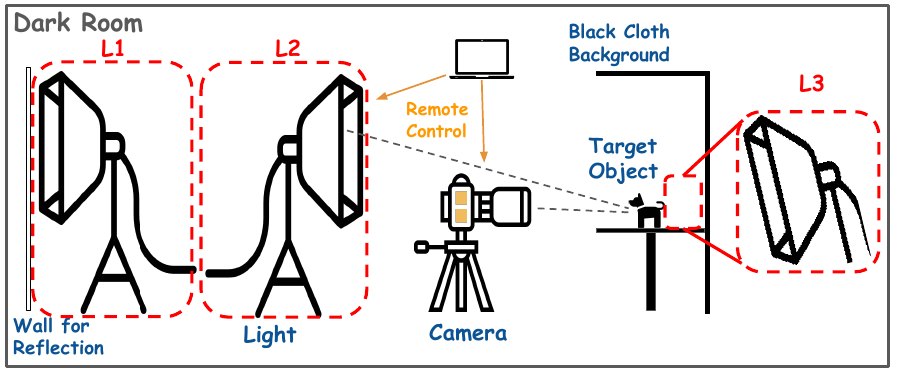}
    \caption{Physical acquisition setup for the real-object pilot study.}
    \label{fig:physical_setup2}
\end{figure}

\subsection{Real-Object Pilot Study}
The previous four benchmarks, despite using different acquisition protocols, all rely on recapturing printed images. Therefore, the open question is whether adaptive sensor control still has an advantage over conventional baselines when capturing real-world objects. To examine performance without printed-image recapture, we conduct a real-object pilot study in which physical objects are directly captured under varying illumination conditions and camera parameter configurations.

The physical setup follows the general layout in Figure~\ref{fig:physical_setup}, but replaces the printed target with a physical object. We cover both the tabletop and background with black cloth to minimize visual distractions, provide a clean background, and suppress uncontrolled reflections. We deliberately select scaled replicas of real-world objects and place each object at a fixed position in the scene. We select four physical object samples spanning three semantic categories: one dog, two cars, and one frog.

\begin{table}[t]
\centering

\begin{minipage}[t]{0.36\linewidth}
\vspace{0pt}
\centering
\caption{Lighting conditions.}
\label{tab:real_object_lighting}
\small
\setlength{\tabcolsep}{4pt}
\begin{tabular}{@{}ccc@{}}
    \toprule
    \textbf{ID} &
    \shortstack{\textbf{Type}} &
    \textbf{Intensity (\%)} \\
    \midrule
    L1 & Reflected & 1   \\
    L2 & Front-lit & 30  \\
    L3 & Backlit   & 100 \\
    \bottomrule
\end{tabular}
\end{minipage}\hfill
\begin{minipage}[t]{0.60\linewidth}
\vspace{0pt}
\centering
\caption{Manual settings for the real-object pilot study.}
\label{tab:real_object_exposure}
\small
\setlength{\tabcolsep}{4pt}
\begin{tabularx}{\linewidth}{
    @{}l>{\raggedright\arraybackslash}X@{}
}
    \toprule
    \textbf{Parameter} & \textbf{Values} \\
    \midrule
    Aperture &
    $f$/2.8, $f$/4, $f$/8, $f$/11, $f$/16, $f$/22 \\
    \addlinespace[2pt]
    ISO &
    100, 250, 800, 2000, 3200, 6400, 12800, 32000 \\
    \addlinespace[2pt]
    \shortstack[l]{Shutter speed} &
    1/2\,s, 1/3\,s, 1/15\,s, 1/60\,s,
    1/250\,s, 1/1000\,s \\
    \bottomrule
\end{tabularx}
\end{minipage}

\end{table}
To evaluate sensor control under more challenging illumination, we deliberately vary both the light position and intensity. Figure~\ref{fig:physical_setup2} illustrates the resulting lighting geometries, while Table~\ref{tab:real_object_lighting} summarizes the corresponding light positions and intensity settings. For each object--illumination combination, we further capture images under different exposure settings, as summarized in Table~\ref{tab:real_object_exposure}.

For each object, we define a square crop using a reference AE image and reuse the same crop coordinates across all manual exposure settings and lighting conditions. Both AE and manual-exposure crops are resized directly to $576\times576$ using Lanczos resampling. Additional details of the real-object pilot study are provided in Appendix~\ref{Appendix: real-object pipeline}.

\section{Results}
\paragraph{Evaluation protocol.}
Our primary goal is to examine whether conclusions about adaptive sensor control remain consistent across different benchmark acquisition pipelines. Within each pipeline, we evaluate exposure-control strategies using captured images of each sample under each lighting condition, following the evaluation task of Lens~\cite{baek2025adaptive}. For the printed-image benchmarks, we evaluate 200-way classification using 12 downstream model configurations covering the 11 models considered in Lens. The same 12 configurations are used in the real-object pilot study.

We compared against the following baselines: \textit{AE (Auto Exposure)} uses the camera’s built-in exposure control, with accuracy averaged over all shots for each sample and lighting condition. \textit{Lens} selects an exposure setting based on downstream confidence for each model and sample. \textit{Random} averages accuracy over all manual exposure candidates. \textit{Oracle-S} (Oracle-Specific) selects the best exposure for each sample and model, providing an upper bound on sample-specific control. \textit{Oracle-F} (Oracle-Fixed) retrospectively selects the manual exposure setting that maximizes average classification accuracy across all evaluated model configurations, samples, and lighting conditions within each pipeline, and uses that setting throughout the pipeline. Unless otherwise stated, we report top-1 classification accuracy averaged across lighting conditions.

\subsection{Controlled Pipeline Results}

\subsubsection{Quantitative Results}
Table~\ref{tab:pipeline_overall_results} compares the average classification accuracy across all 12 downstream model configurations under our four controlled acquisition pipelines and the original ImageNet-ES-Diverse (IED) benchmark. For IED, we retrieve the corresponding 20 source samples from the released benchmark and evaluate them using the same 12 downstream model configurations.


\begin{table}[t]
    \centering
    \caption{Average downstream classification accuracy (\%) under different acquisition pipelines, averaged across all 12 downstream model configurations. IED denotes ImageNet-ES-Diverse.}
    \label{tab:pipeline_overall_results}
    \small
    \setlength{\tabcolsep}{4pt}
    \begin{tabular}{llccccc}
        \toprule
        \textbf{Pipeline} &
        \textbf{Print / framing} &
        \textbf{AE} &
        \textbf{Lens} &
        \textbf{Oracle-S} &
        \textbf{Oracle-F} &
        \textbf{Random} \\
        \midrule
        a & Original pixels / 28\,mm
          & 53.89 & 52.33 & 79.17 & 55.17 & 31.18 \\
        b & Original pixels / 70\,mm
          & 60.94 & 59.50 & 81.33 & 60.50 & 37.35 \\
        c & Fixed-width print / 28\,mm
          & 86.31 & 85.08 & 94.67 & 86.67 & 56.43 \\
        d & Fixed-width print / zoomed
          & 87.61 & 87.00 & 95.75 & 87.25 & 58.86 \\
        \midrule
        IED & Released benchmark
            & 28.43 & 49.03 & 74.58 & 51.46 & 19.79 \\
        \bottomrule
    \end{tabular}
\end{table}
The results reveal substantial variation in downstream accuracy across acquisition pipelines. In particular, pipelines (c) and (d), which use a larger physically printed source image, improve accuracy on all baselines over pipeline (a). Increasing camera zoom alone also improves performance, as shown by the difference between pipelines (a)/(b) and (c)/(d). More importantly, the relative performance between AE and Lens also changes across benchmark constructions. On average, Lens substantially outperforms AE on IED, whereas AE slightly outperforms Lens under all four of our independently collected pipelines. We provide more analysis in Appendix~\ref{Appendix:Quan_analysis}.

\subsubsection{Qualitative Analysis}
\begin{figure*}[t]
\centering
\includegraphics[width=\textwidth]{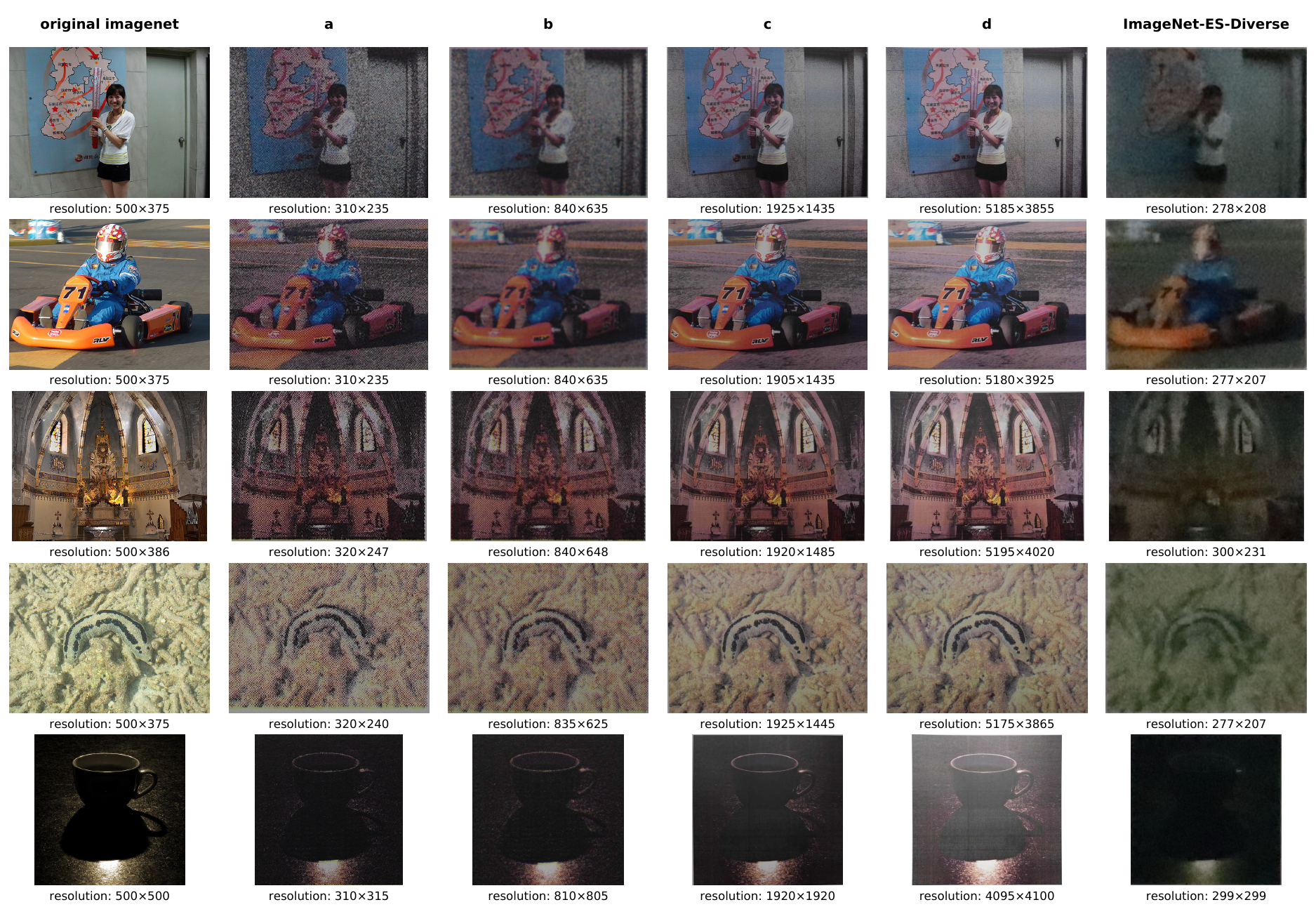}
\caption{Representative source images and their corresponding Auto Exposure (AE) captures under ImageNet-ES-Diverse (IED) and our four acquisition pipelines (a)--(d). Each row shows the same source sample across all benchmarks. Our captures use 50\% light intensity, while IED captures use L2; these lighting conditions are not physically matched.}
\label{fig:AE_vis}
\end{figure*}

To visualize how the acquisition pipeline changes the resulting observations, Figure~\ref{fig:AE_vis} shows the first five source samples in our 20-sample evaluation set together with their corresponding AE captures under IED and pipelines (a)--(d). For our four pipelines, we show AE captures obtained at 50\% light intensity; for IED, we retrieve the released AE captures under the L2 lighting condition. 

The visual differences are substantial, particularly between ImageNet-ES-Diverse and the original source image. The IED AE captures are noticeably darker and blurrier, with substantially less visible object detail than other pipelines. Among our controlled pipelines, (a) and (b), which use the smaller physically printed source images, also exhibit reduced fine detail and more visible printing artifacts, whereas (c) and (d), which use larger printed images, preserve substantially clearer object structure after recapture. Increasing camera zoom provides an additional improvement in effective target scale.


\subsection{Real-Object Pilot Results}

Table~\ref{tab:real_object_results} reports per-model classification accuracy in the real-object pilot study. The downstream models are configured to predict over the 1000-way ImageNet-1K classes. For each of the three physical object categories, we define a corresponding subset of fine-grained ImageNet-1K classes. A prediction is considered correct if the model's top-1 predicted class belongs to the subset associated with the ground-truth object category.

\begin{table*}[t]
\centering
\caption{Classification accuracy (\%) in the real-object pilot study. DA: DeepAugment; AM: AugMix.}
\label{tab:real_object_results}
\small
\setlength{\tabcolsep}{5pt}
\begin{tabular}{@{}lccccc@{}}
\toprule
\textbf{Model} & \textbf{AE} & \textbf{Lens} &
\textbf{Oracle-S} & \textbf{Oracle-F} & \textbf{Random} \\
\midrule
ResNet-50          & 50.00 & 58.33 & 100.00 & 33.33 & 26.68 \\
ResNet-50 (DA+AM)  & 66.67 & 66.67 & 91.67  & 50.00 & 42.85 \\
ResNet-152         & 58.33 & 41.67 & 91.67  & 33.33 & 33.83 \\
EfficientNet-B0    & 66.67 & 66.67 & 91.67  & 33.33 & 35.85 \\
EfficientNet-B3    & 66.67 & 50.00 & 100.00 & 50.00 & 45.08 \\
SwinV2-T           & 58.33 & 41.67 & 100.00 & 58.33 & 39.15 \\
SwinV2-S           & 75.00 & 50.00 & 100.00 & 66.67 & 42.77 \\
SwinV2-B           & 41.67 & 25.00 & 100.00 & 58.33 & 40.10 \\
OpenCLIP-b         & 50.00 & 58.33 & 100.00 & 66.67 & 42.91 \\
OpenCLIP-h         & 91.67 & 75.00 & 100.00 & 75.00 & 55.47 \\
DINOv2-b           & 58.33 & 41.67 & 100.00 & 58.33 & 41.96 \\
DINOv2-g           & 66.67 & 41.67 & 100.00 & 75.00 & 48.96 \\
\midrule
\textbf{Average} & \textbf{62.50} & \textbf{51.39} &
\textbf{97.92} & \textbf{54.86} & \textbf{41.30} \\
\bottomrule
\end{tabular}
\end{table*}
The overall trend is consistent with our collected printed-image pipelines. Across the reported model configurations,  AE outperforms Lens for most configurations, while Lens performs better on ResNet-50 and OpenCLIP-b. Thus, the substantial Lens-over-AE advantage observed on ImageNet-ES-Diverse does not persist in our real-object pilot study. This provides preliminary evidence that the observed adaptive-control advantage may not extend to other physical acquisition settings.


\section{Discussion}

Our results show that the acquisition protocol of a physical benchmark affects the downstream task difficulty and the measured effectiveness of adaptive sensor-control strategies. The substantial Lens-over-AE advantage observed on IED does not persist on average across our four pipelines or real-object pilot. Therefore, conclusions drawn under a single acquisition protocol may not generalize to other physical settings. Our real-object pilot study provides preliminary evidence suggesting that the adaptive advantage observed on a particular recapture-based benchmark should not be interpreted as a general property of physical sensing.

On IED, Lens approaches Oracle-F but remains well below Oracle-S, whereas AE often approaches Oracle-F in our pipelines despite substantial Oracle-S headroom. This suggests that AE quality, adaptive headroom, and controller capability should be considered separately. 

Future physical benchmarks should distinguish intended environmental and sensor variations from protocol-induced effects and explicitly report acquisition details. A useful next step is to directly capture real scenes under multiple controlled environmental conditions and sensor configurations to test whether conclusions from recapture-based benchmarks persist in more realistic settings.

The full IED recapture protocol is not publicly available, so our reference setup relies on published descriptions and released dataset statistics. Access to the original printable PDFs and detailed metadata would help identify factors behind the observed benchmark shift. In addition, our real-object pilot covers only four samples under three lighting conditions. While providing preliminary support for the recapture results, a larger dataset is needed to assess generalization to real-world settings.

\section{Acknowledgement}
The research reported in this paper was sponsored in part by the DEVCOM Army Research Laboratory (award \#W911NF1720196), National Institutes of Health (award \#1P41EB028242), and National Science Foundation (award \#CNS-2325956). The views and
conclusions contained in this document are those of the author(s) and should
not be interpreted as representing the official policies of the funding agencies.

\bibliographystyle{unsrt}
\bibliography{references}

\newpage


\appendix

\section{Recapture Pipelines Comparison}
\subsection{Cropped Resolution}
\label{Appendix:resolution}

Table~\ref{tab:cropped-resolution-20-samples} reports the source-image dimensions and the cropped-image dimensions in ImageNet-ES-Diverse and pipelines (a)--(d) for the same 20 source samples. This provides a per-sample comparison of image dimensions across the datasets.

\begin{table*}[t]
\centering
\caption{Image resolutions ($W\times H$, pixels) for all 20 samples in the source images, ImageNet-ES-Diverse, and four acquisition pipelines. The four local capture columns report ROI crop sizes; Original reports the original ImageNet size, and ES-Diverse reports the provided image size at \texttt{l2/param\_2}. The first five rows follow the sample order in Figure~\ref{fig:AE_vis}. Sample IDs omit the shared \texttt{ILSVRC2012\_val\_} prefix.}
\label{tab:cropped-resolution-20-samples}
\setlength{\tabcolsep}{5pt}
\small
\resizebox{\textwidth}{!}{%
\begin{tabular}{lcccccc}
\toprule
Sample ID & Original & ES-Diverse & a & b & c & d \\
\midrule
\texttt{00002331} & $500\times 375$ & $278\times 208$ & $310\times 235$ & $840\times 635$ & $1925\times 1435$ & $5185\times 3855$ \\
\texttt{00005381} & $500\times 375$ & $277\times 207$ & $310\times 235$ & $840\times 635$ & $1905\times 1435$ & $5180\times 3925$ \\
\texttt{00011526} & $500\times 386$ & $300\times 231$ & $320\times 247$ & $840\times 648$ & $1920\times 1485$ & $5195\times 4020$ \\
\texttt{00011857} & $500\times 375$ & $277\times 207$ & $320\times 240$ & $835\times 625$ & $1925\times 1445$ & $5175\times 3865$ \\
\texttt{00015548} & $500\times 500$ & $299\times 299$ & $310\times 315$ & $810\times 805$ & $1920\times 1920$ & $4095\times 4100$ \\
\texttt{00038504} & $500\times 375$ & $276\times 207$ & $320\times 240$ & $835\times 625$ & $1980\times 1495$ & $5230\times 3895$ \\
\texttt{00034232} & $500\times 375$ & $272\times 204$ & $320\times 240$ & $845\times 635$ & $1985\times 1505$ & $5260\times 3875$ \\
\texttt{00048338} & $383\times 500$ & $209\times 272$ & $245\times 320$ & $645\times 840$ & $1930\times 2520$ & $2985\times 3915$ \\
\texttt{00023874} & $500\times 375$ & $281\times 210$ & $300\times 225$ & $805\times 600$ & $1900\times 1450$ & $5190\times 3890$ \\
\texttt{00028145} & $500\times 375$ & $283\times 212$ & $305\times 230$ & $860\times 645$ & $1890\times 1420$ & $5215\times 3870$ \\
\texttt{00031428} & $500\times 500$ & $300\times 300$ & $310\times 310$ & $805\times 805$ & $1940\times 1925$ & $4205\times 4245$ \\
\texttt{00030373} & $500\times 500$ & $302\times 302$ & $310\times 315$ & $835\times 845$ & $1920\times 1915$ & $4120\times 4075$ \\
\texttt{00048182} & $500\times 500$ & $307\times 307$ & $310\times 305$ & $835\times 830$ & $1900\times 1940$ & $4155\times 4105$ \\
\texttt{00043088} & $500\times 458$ & $300\times 274$ & $310\times 285$ & $810\times 742$ & $1890\times 1710$ & $4440\times 4055$ \\
\texttt{00020614} & $500\times 500$ & $300\times 300$ & $310\times 310$ & $870\times 870$ & $1890\times 1915$ & $4150\times 4130$ \\
\texttt{00016634} & $500\times 480$ & $297\times 285$ & $315\times 307$ & $832\times 807$ & $1920\times 1848$ & $4060\times 3910$ \\
\texttt{00012502} & $2304\times 1728$ & $1370\times 1027$ & $1440\times 1083$ & $3848\times 2893$ & $1925\times 1445$ & $5200\times 3885$ \\
\texttt{00016500} & $500\times 375$ & $283\times 212$ & $320\times 240$ & $835\times 625$ & $1895\times 1430$ & $5085\times 3815$ \\
\texttt{00036886} & $459\times 765$ & $243\times 405$ & $285\times 480$ & $760\times 1277$ & $1886\times 3150$ & $2491\times 4095$ \\
\texttt{00027894} & $500\times 375$ & $279\times 209$ & $318\times 232$ & $835\times 625$ & $1880\times 1435$ & $5000\times 3747$ \\
\bottomrule
\end{tabular}%
}
\end{table*}

\subsection{Per-configuration Accuracy on Four Pipelines and IED}
\label{Appendix:per-model accuracy}
Tables~\ref{tab:20samples_a},
\ref{tab:20samples_b},
\ref{tab:20samples_c},
\ref{tab:20samples_d}, and
\ref{tab:20samples_ied}
report the top-1 classification accuracy of each of the 12 downstream model configurations under
pipelines (a)--(d) and ImageNet-ES-Diverse, respectively.

\begin{table*}[t]
\centering
\caption{Classification accuracy (\%) for Pipeline a. DA: DeepAugment; AM: AugMix.}
\label{tab:20samples_a}
\setlength{\tabcolsep}{5pt}
\begin{tabular}{@{}lccccc@{}}
\toprule
\textbf{Model} & \textbf{AE} & \textbf{Lens} &
\textbf{Oracle-S} & \textbf{Oracle-F} & \textbf{Random} \\
\midrule
ResNet-50 & 20.00 & 33.00 & 60.00 & 25.00 & 14.81 \\
ResNet-50 (DA+AM) & 64.33 & 53.00 & 87.00 & 58.00 & 33.11 \\
ResNet-152 & 33.67 & 34.00 & 72.00 & 34.00 & 18.11 \\
EfficientNet-B0 & 26.00 & 35.00 & 58.00 & 31.00 & 19.19 \\
EfficientNet-B3 & 54.33 & 44.00 & 79.00 & 61.00 & 29.70 \\
SwinV2-T & 42.33 & 39.00 & 66.00 & 44.00 & 22.44 \\
SwinV2-S & 49.33 & 49.00 & 79.00 & 52.00 & 26.44 \\
SwinV2-B & 44.00 & 43.00 & 78.00 & 43.00 & 26.67 \\
OpenCLIP-b & 59.67 & 60.00 & 82.00 & 63.00 & 38.00 \\
OpenCLIP-h & 80.67 & 78.00 & 94.00 & 84.00 & 47.44 \\
DINOv2-b & 83.33 & 76.00 & 95.00 & 83.00 & 45.81 \\
DINOv2-g & 89.00 & 84.00 & 100.00 & 84.00 & 52.41 \\
\midrule
\textbf{Average} & \textbf{53.89} & \textbf{52.33} & \textbf{79.17} & \textbf{55.17} & \textbf{31.18} \\
\bottomrule
\end{tabular}
\end{table*}

\begin{table*}[t]
\centering
\caption{Classification accuracy (\%) for Pipeline b. DA: DeepAugment; AM: AugMix.}
\label{tab:20samples_b}
\setlength{\tabcolsep}{5pt}
\begin{tabular}{@{}lccccc@{}}
\toprule
\textbf{Model} & \textbf{AE} & \textbf{Lens} &
\textbf{Oracle-S} & \textbf{Oracle-F} & \textbf{Random} \\
\midrule
ResNet-50 & 30.00 & 29.00 & 59.00 & 24.00 & 16.56 \\
ResNet-50 (DA+AM) & 61.00 & 69.00 & 85.00 & 64.00 & 40.33 \\
ResNet-152 & 49.00 & 48.00 & 65.00 & 44.00 & 22.59 \\
EfficientNet-B0 & 33.33 & 37.00 & 67.00 & 29.00 & 20.44 \\
EfficientNet-B3 & 54.00 & 57.00 & 84.00 & 57.00 & 35.70 \\
SwinV2-T & 50.33 & 52.00 & 75.00 & 55.00 & 28.78 \\
SwinV2-S & 60.00 & 52.00 & 81.00 & 59.00 & 32.19 \\
SwinV2-B & 62.33 & 59.00 & 84.00 & 66.00 & 36.67 \\
OpenCLIP-b & 60.33 & 60.00 & 81.00 & 61.00 & 42.59 \\
OpenCLIP-h & 85.00 & 83.00 & 100.00 & 83.00 & 56.48 \\
DINOv2-b & 93.33 & 79.00 & 95.00 & 90.00 & 53.19 \\
DINOv2-g & 92.67 & 89.00 & 100.00 & 94.00 & 62.67 \\
\midrule
\textbf{Average} & \textbf{60.94} & \textbf{59.50} & \textbf{81.33} & \textbf{60.50} & \textbf{37.35} \\
\bottomrule
\end{tabular}
\end{table*}

\begin{table*}[t]
\centering
\caption{Classification accuracy (\%) for Pipeline c. DA: DeepAugment; AM: AugMix.}
\label{tab:20samples_c}
\setlength{\tabcolsep}{5pt}
\begin{tabular}{@{}lccccc@{}}
\toprule
\textbf{Model} & \textbf{AE} & \textbf{Lens} &
\textbf{Oracle-S} & \textbf{Oracle-F} & \textbf{Random} \\
\midrule
ResNet-50 & 70.00 & 73.00 & 85.00 & 73.00 & 39.81 \\
ResNet-50 (DA+AM) & 71.00 & 80.00 & 92.00 & 74.00 & 48.30 \\
ResNet-152 & 86.33 & 89.00 & 95.00 & 85.00 & 46.70 \\
EfficientNet-B0 & 69.67 & 72.00 & 91.00 & 70.00 & 43.89 \\
EfficientNet-B3 & 86.67 & 85.00 & 91.00 & 85.00 & 53.74 \\
SwinV2-T & 85.00 & 78.00 & 92.00 & 86.00 & 57.15 \\
SwinV2-S & 92.00 & 89.00 & 95.00 & 91.00 & 60.44 \\
SwinV2-B & 90.00 & 85.00 & 95.00 & 93.00 & 61.44 \\
OpenCLIP-b & 100.00 & 96.00 & 100.00 & 100.00 & 65.59 \\
OpenCLIP-h & 95.00 & 92.00 & 100.00 & 94.00 & 67.22 \\
DINOv2-b & 95.00 & 92.00 & 100.00 & 96.00 & 65.93 \\
DINOv2-g & 95.00 & 90.00 & 100.00 & 93.00 & 66.93 \\
\midrule
\textbf{Average} & \textbf{86.31} & \textbf{85.08} & \textbf{94.67} & \textbf{86.67} & \textbf{56.43} \\
\bottomrule
\end{tabular}
\end{table*}

\begin{table*}[t]
\centering
\caption{Classification accuracy (\%) for Pipeline d. DA: DeepAugment; AM: AugMix.}
\label{tab:20samples_d}
\setlength{\tabcolsep}{5pt}
\begin{tabular}{@{}lccccc@{}}
\toprule
\textbf{Model} & \textbf{AE} & \textbf{Lens} &
\textbf{Oracle-S} & \textbf{Oracle-F} & \textbf{Random} \\
\midrule
ResNet-50 & 71.33 & 75.00 & 85.00 & 73.00 & 40.96 \\
ResNet-50 (DA+AM) & 71.67 & 80.00 & 88.00 & 74.00 & 49.52 \\
ResNet-152 & 84.67 & 88.00 & 95.00 & 85.00 & 48.48 \\
EfficientNet-B0 & 76.33 & 80.00 & 92.00 & 73.00 & 46.52 \\
EfficientNet-B3 & 90.00 & 85.00 & 95.00 & 87.00 & 56.89 \\
SwinV2-T & 85.00 & 79.00 & 95.00 & 85.00 & 59.59 \\
SwinV2-S & 92.67 & 87.00 & 100.00 & 94.00 & 63.74 \\
SwinV2-B & 95.00 & 89.00 & 99.00 & 95.00 & 65.93 \\
OpenCLIP-b & 100.00 & 100.00 & 100.00 & 100.00 & 69.67 \\
OpenCLIP-h & 95.00 & 94.00 & 100.00 & 94.00 & 70.19 \\
DINOv2-b & 94.67 & 96.00 & 100.00 & 93.00 & 66.22 \\
DINOv2-g & 95.00 & 91.00 & 100.00 & 94.00 & 68.63 \\
\midrule
\textbf{Average} & \textbf{87.61} & \textbf{87.00} & \textbf{95.75} & \textbf{87.25} & \textbf{58.86} \\
\bottomrule
\end{tabular}
\end{table*}

\begin{table*}[t]
\centering
\caption{Classification accuracy (\%) on ImageNet-ES-Diverse. DA: DeepAugment; AM: AugMix.}
\label{tab:20samples_ied}

\setlength{\tabcolsep}{5pt}
\begin{tabular}{@{}lccccc@{}}
\toprule
\textbf{Model} & \textbf{AE} & \textbf{Lens} &
\textbf{Oracle-S} & \textbf{Oracle-F} & \textbf{Random} \\
\midrule
ResNet-50 & 11.67 & 34.17 & 55.83 & 32.50 & 10.19 \\
ResNet-50 (DA+AM) & 26.00 & 57.50 & 79.17 & 55.83 & 20.99 \\
ResNet-152 & 9.00 & 42.50 & 64.17 & 40.83 & 12.69 \\
EfficientNet-B0 & 19.67 & 35.00 & 64.17 & 37.50 & 14.23 \\
EfficientNet-B3 & 28.67 & 38.33 & 70.83 & 50.00 & 17.56 \\
SwinV2-T & 21.50 & 40.83 & 69.17 & 44.17 & 16.88 \\
SwinV2-S & 24.67 & 43.33 & 72.50 & 40.00 & 16.05 \\
SwinV2-B & 29.67 & 35.00 & 63.33 & 40.83 & 16.02 \\
OpenCLIP-b & 37.83 & 60.00 & 80.83 & 60.83 & 24.88 \\
OpenCLIP-h & 42.67 & 72.50 & 85.83 & 75.83 & 30.99 \\
DINOv2-b & 36.50 & 62.50 & 89.17 & 70.00 & 25.99 \\
DINOv2-g & 53.33 & 66.67 & 100.00 & 69.17 & 31.05 \\
\midrule
\textbf{Average} & \textbf{28.43} & \textbf{49.03} & \textbf{74.58} & \textbf{51.46} & \textbf{19.79} \\
\bottomrule
\end{tabular}
\end{table*}

\subsection{Supplementary Analysis on Four-Pipeline Results}
\label{Appendix:Quan_analysis}

An additional observation is that, although pipeline (a) is designed as our reference recapture setup informed by ImageNet-ES-Diverse, the Lens-over-AE advantage observed on the released ImageNet-ES-Diverse subset does not persist in pipeline (a). This discrepancy cannot be attributed to a single factor from our current experiments. Based on qualitative inspection of the released dataset, the reported acquisition setup, and our own recapture results, we identify four plausible hypotheses:

\begin{itemize}
\item \textbf{Printing and recapture quality.} Compared with both the original ImageNet images and our recaptured samples, the released ImageNet-ES-Diverse crops visually exhibit stronger loss of fine details and texture. Such degradation may originate from differences in source-to-print scaling, printing quality, optical sampling, focus, or subsequent cropping and resizing. These factors could affect both absolute classification difficulty and the confidence ranking used by Lens.

\item \textbf{Metering area.} The AE captures in ImageNet-ES-Diverse often appear relatively dark within the cropped target region. One possible explanation is a mismatch between the region used by the camera for exposure metering and the region ultimately provided to the classifier. If the printed target occupies only a small portion of a frame containing a large white page or background, multi-area metering may select an exposure appropriate for the full frame but underexpose the cropped ROI. This mechanism is plausible but cannot be verified without the original uncropped captures and corresponding exposure metadata.

\item \textbf{Lighting conditions.} Differences in illumination remain another possible source of the discrepancy. However, based on the reported acquisition setup, ImageNet-ES-Diverse does not appear to deliberately introduce extreme lighting geometries intended to challenge AE: the setup uses two Philips lights near the target and varies their intensity. Our reference setup similarly places a studio light approximately in front of the sample and varies its intensity. Therefore, although the two lighting setups are not physically matched and illumination differences cannot be excluded, we do not find clear evidence that substantially more difficult lighting alone explains the large Lens--AE discrepancy.

\item \textbf{Camera and lens differences.} ImageNet-ES-Diverse and our experiments use different camera bodies and lenses, and camera-dependent metering, Auto ISO policies, tone mapping, and JPEG processing may affect AE behavior. Although both setups use modern mirrorless camera systems, we cannot assume that their AE implementations behave equivalently. Hardware- and firmware-dependent imaging pipelines therefore remain a possible confounding factor that is not isolated in our current experiments.

\end{itemize}

These hypotheses are intended to identify possible sources of the cross-dataset discrepancy rather than to establish a causal explanation. Isolating their individual effects would require access to the original uncropped captures and metadata or controlled experiments that independently vary printing quality, metering configuration, illumination, and camera hardware.

\section{Real-Object Pilot Study Details}
\label{Appendix: real-object pipeline}
\subsection{Camera Parameters}
Table~\ref{tab:real_object_camera_settings} is a summary of camera settings used during the capture.


\begin{table}[t]
\centering
\caption{Camera settings for the real-object pilot study.}
\label{tab:real_object_camera_settings}
\small
\begin{tabular}{@{}lp{0.57\linewidth}@{}}
\toprule
\textbf{Setting} & \textbf{Value} \\
\midrule
Camera & Sony ILCE-7RM6 \\
Lens & Sony FE 24--70mm F2.8 GM II \\
Focal length & 35\,mm \\
AE mode & Program Auto (P) \\
White balance & Auto (AWB) \\
Metering & Pattern (multi-segment) \\
Image format & JPEG \\
Original resolution & $3328 \times 3328$ pixels \\
Focus mode & Manual focus (MF) \\
Drive mode & Single shooting \\
Creative Look & Standard \\
\bottomrule
\end{tabular}
\end{table}

\subsection{Class Mapping}
Table~\ref{tab:real_object_class_mapping} maps the three object categories in the pilot study to fine-grained ImageNet classes.
\begin{table*}[t]
\centering
\caption{Mapping of the three real-object classes to ImageNet-1K subclasses.
Indices are zero-based; ranges include both endpoints.}
\label{tab:real_object_class_mapping}
\small
\setlength{\tabcolsep}{5pt}
\begin{tabular}{@{}clcp{0.62\textwidth}@{}}
\toprule
\textbf{Class ID} & \textbf{Class} & \textbf{Count} &
\textbf{ImageNet subclass indices} \\
\midrule
0 & Dog  & 118 & 151--268 \\
1 & Car  & 12  & 436, 468, 511, 575, 609, 627, 656, 717, 734, 751, 757, 817 \\
2 & Frog & 3   & 30--32 \\
\bottomrule
\end{tabular}
\end{table*}

\subsection{Post-processing: Cropping}

We applied object-centered square cropping to the
$3328\times3328$ captured images to reduce the surrounding
background. Candidate object regions were generated using SAM
(ViT-B) and selected manually. The selected bounding boxes were
mapped to the original image coordinates and expanded into square crops with surrounding context, with crop positions adjusted to remain within the image boundaries. For each object, one square crop was defined using a reference AE image. The same crop coordinates were applied to the AE images and all manual captures across the 288 exposure settings and three lighting conditions. All crops were resized directly to $576\times576$ using Lanczos resampling. Table~\ref{tab:real_object_crop_resolutions} summarizes the crop resolutions for all four evaluated objects.
These resolutions precede classifier input preprocessing.

\begin{table*}[t]
\centering
\caption{Crop resolutions ($W\times H$, pixels) for the four real-object samples. All source images are $3328\times3328$. For each object, the same crop coordinates are shared by AE and manual captures across all lighting conditions and manual exposure settings. Crop sizes are reported before resizing; all crops are resized directly to $576\times576$.}
\label{tab:real_object_crop_resolutions}
\small
\setlength{\tabcolsep}{5pt}
\begin{tabular}{@{}lcc@{}}
\toprule
\textbf{Sample} & \textbf{Shared crop size} & \textbf{Final size} \\
\midrule
Dog   & $575\times575$   & $576\times576$ \\
Car 1 & $2391\times2391$ & $576\times576$ \\
Car 2 & $1858\times1858$ & $576\times576$ \\
Frog  & $658\times658$   & $576\times576$ \\
\bottomrule
\end{tabular}
\end{table*}
\subsection{Visualization}
\begin{figure}[t]
    \centering
    \includegraphics[width=\linewidth]{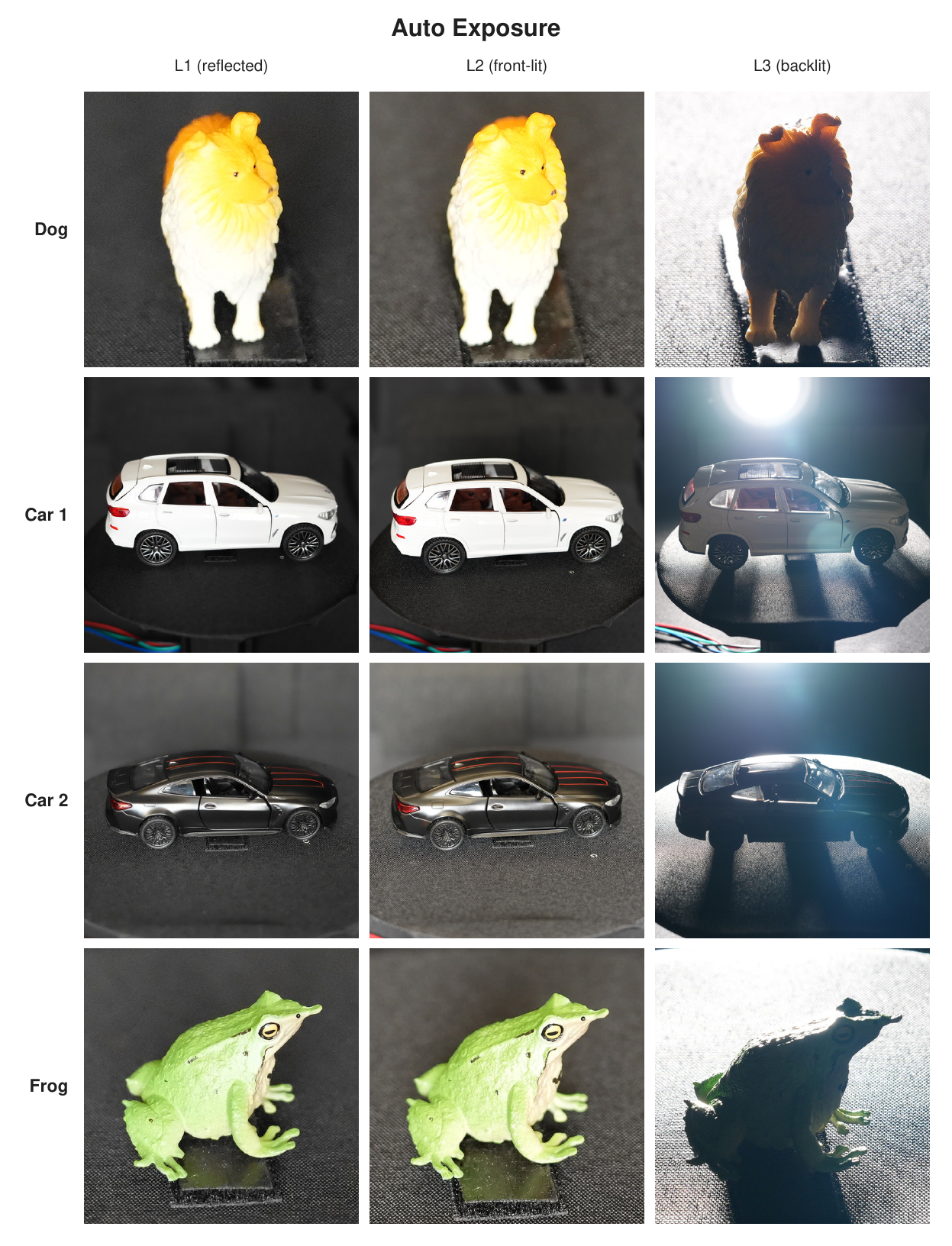}
    \caption{Real-object samples captured using auto exposure (AE) under the three lighting conditions: reflected, front-lit, and backlit.}
    \label{fig:real_object_vis}
\end{figure}

Figure~\ref{fig:real_object_vis} shows the AE samples under the three lighting conditions, processed using the shared cropping and resizing procedure described above.

\end{document}